# BioVid：Autoregressive Video Generation with Biological Behavior Semantic Comprehension

Tsung-Wei Pan[a], Jung-Hua Wang[b,*]
[a]Department of Electrical Engineering, National Taiwan Ocean University, Keelung, Taiwan
[b]AI research center, National Taiwan Ocean University, Keelung, Taiwan

## Abstract

Video generation for biological behavior requires more than visually plausible motion: the duration of an action is itself a semantic property. Existing models usually rely on fixed temporal windows, external continuation, or prompt-driven stories, so length is specified externally rather than learned from behavior. To address this gap, we propose BioVid, a data-driven autoregressive framework for adaptive-length biological behavior generation. BioVid uses a 2D-encode/3D-decode tokenizer: a two-dimensional FSQ-R3GAN encoder converts each frame into discrete visual tokens, preserving single-frame information suited for next-token prediction and EOS-based termination, while a temporally inflated and video-finetuned three-dimensional decoder reconstructs generated tokens with temporal context to reduce flickering. A causal Transformer then models the frame-wise token sequence and, conditioned only on the first frame, stops generation when it emits an End-of-Sequence token, allowing duration to emerge from the learned behavior distribution. We evaluate BioVid on the A001 drinking action from NTU RGB+D. On 94 held-out clips, BioVid achieves a Wasserstein-1 distance of 1.24 frames from the real duration distribution. In comparison, fixed-length baselines yield distances of approximately 6–7 frames even when configured to the available length closest to the dataset mean, and approximately 15 frames when using the conventional 16-frame generation length. These results demonstrate BioVid's ability to learn and reproduce the intrinsic duration distribution of biological behavior.

## 1 Introduction

Biological behavior video generation is not merely a problem of synthesizing visually plausible motion. For behavioral data, the duration of an event is itself part of the semantics: an action ends when the underlying behavior reaches completion, not when an externally specified number of frames has been generated. A drinking action, for example, is complete only after the subject reaches for the cup, brings it to the mouth, performs the drinking motion, and returns toward a terminal posture. The number of frames required for this process varies across subjects, body scales, motion speeds, camera views, and execution styles. Therefore, the temporal boundary of a behavioral event should not be treated as a nuisance variable to be manually prescribed, but as a statistical property of the behavior itself.

This property is especially important in data-driven biological research. Unlike general-purpose video generation, where users may specify a desired duration through a text prompt or interface, biological behavior synthesis often aims to model the empirical structure of observed behaviors. In many scientific settings, the target behavior may be difficult to describe precisely in language, particularly for rare species, non-human organisms, or subtle physiological actions. In such cases, observational data provide a more reliable source of behavioral semantics than human-authored descriptions. A model for biological behavior generation should therefore learn not only the appearance and motion patterns of a behavior, but also its natural duration distribution.

Recent diffusion-based video models have achieved remarkable progress in visual fidelity, temporal coherence, and prompt-based controllability. Models such as Video Diffusion Models, Imagen Video, Make-A-Video, Video LDM, and DiT-based systems demonstrate the effectiveness of denoising-based generation for high-quality text-to-video and latent video synthesis [1], [2], [3], [4], [5]. However, these models generally synthesize a continuous spatio-temporal representation over a prescribed temporal window. Their output duration is therefore determined before sampling begins, and the continuous representation does not provide a natural categorical position at which a semantic stop signal can be predicted. As a result, diffusion-based models are well suited to generating high-quality videos of a specified duration, but they do not directly address how a model should decide that a behavioral event has naturally ended.

Autoregressive and token-based video models provide a more relevant foundation for adaptive temporal generation. Models such as VideoGPT, HARP, TATS, MAGVIT, and VideoPoet show that videos can be represented as latent visual tokens and modeled with Transformer architectures for future prediction, long-form continuation, masked generation, or multimodal synthesis [6], [7], [8], [9], [10]. However, their output duration is still typically governed by fixed clip formats, task definitions, rollout procedures, or prompt structures. They address how to predict, reconstruct, continue, or control video content, but not how to learn from data when a specific behavior should terminate.

A small number of recent models appear closer to adaptive-length generation, but they address a different notion of adaptivity. Phenaki supports variable-length video generation from open-domain textual descriptions by using a causal video tokenizer and time-varying text prompts [11]. TV2TV further decomposes generation into interleaved textual planning and video synthesis, allowing the model to alternate between generating language and video content [12]. These systems improve long-form generation, controllability, and prompt-driven temporal structure, but their adaptivity is driven by external prompts, stories, planning steps, or modality transitions. They do not learn from visual behavior alone when an action has reached semantic completion.

In this work, adaptive-length generation does not mean simply generating videos of arbitrary duration, extending a clip beyond the training horizon, or allowing the user to specify a longer story. Instead, we define adaptive-length behavioral generation as the ability of a model to determine its own termination point from the generated visual sequence, such that the resulting duration distribution matches the natural duration distribution of real behavioral events. Under this definition, long video continuation and semantic termination are different problems. A model may be able to generate longer videos, extrapolate beyond the training length, or follow changing prompts, while still lacking a mechanism for deciding when a particular behavior is complete.

We argue that data-driven semantic termination requires two architectural ingredients. First, the model must operate over a discrete visual representation, so that a stop signal can be represented as a categorical symbol and learned with the same next-token objective as visual content. Second, generation must be temporally autoregressive, so that the decision to continue or terminate can be conditioned on the visual history generated so far. Discrete tokenization without temporal autoregression still requires an externally specified generation length, while temporal autoregression in a continuous latent space lacks a natural symbolic endpoint. Only the combination of discrete visual tokens and causal sequence modeling provides a direct mechanism for learning a behavioral end-of-sequence distribution from data.

To address this gap, we propose BioVid, a data-driven autoregressive framework for biological behavior video generation. BioVid consists of two stages. In the first stage, videos are represented using a 2D-encode/3D-decode FSQ-R3GAN tokenizer. A two-dimensional FSQ-R3GAN encoder maps each video frame into a compact grid of discrete visual tokens. This frame-wise representation preserves single-frame visual information in a form well suited for autoregressive modeling: each group of visual tokens corresponds to one frame, allowing an End-of-Sequence token to be predicted at any frame boundary. To improve video consistency without changing the autoregressive token sequence, reconstruction is performed with a MAGVIT-style temporally inflated and video-finetuned three-dimensional decoder, which uses temporal context to reduce flickering during decoding while preserving frame-level temporal granularity. In the second stage, a causal Transformer models the resulting frame-wise visual token sequence autoregressively. During generation, the model is conditioned only on the first frame and continues predicting visual tokens until it emits the EOS token. Thus, the output duration is not specified by the user, but emerges from the learned distribution over behavioral sequences.

We evaluate BioVid on the NTU RGB+D dataset, a large-scale benchmark of human actions containing RGB, depth, infrared, and skeleton modalities across diverse subjects and viewpoints [13]. This dataset is particularly suitable for studying behavioral termination because its clips correspond to meaningful action instances with empirical temporal boundaries. We focus on the A001 drinking action, whose duration exhibits substantial natural variation across subjects and executions. This setting provides a controlled test case for evaluating whether a model can reproduce not only the visual appearance of an action, but also its duration distribution.

Our experiments show that explicit semantic termination is essential for adaptive-length biological behavior generation. On 94 held-out drinking behavior clips, BioVid produces a generated length distribution that closely matches the test distribution, achieving a Wasserstein-1 distance of 1.24 frames against the ground truth. In contrast, fixed-length baselines manually set near the dataset mean obtain substantially larger distances of approximately 6–7 frames, while the conventional 16-frame setting deviates even further with a distance of 15.48 frames. These results show that matching the average duration alone is insufficient: a model must reproduce the full distribution of behavioral lengths. BioVid achieves this by learning when to stop from data, demonstrating that semantic termination can be modeled as part of visual autoregression.

The contributions of this work are threefold. First, we introduce a data-driven semantic termination mechanism for biological behavior video generation, in which video length is learned as an intrinsic statistical property of the target behavior rather than prescribed as an external parameter. Second, we propose a 2D-encode/3D-decode tokenizer design for autoregressive behavioral video generation. The two-dimensional encoder preserves frame-wise visual tokens that are naturally compatible with next-token prediction and EOS-based termination, while the temporally inflated three-dimensional decoder reduces flickering and improves video consistency without altering the generated token sequence. Third, we provide an empirical evaluation on NTU RGB+D showing that BioVid more faithfully reproduces the natural length distribution of human drinking behavior than fixed-length baselines, establishing adaptive-length generation as a distinct requirement for biologically meaningful video synthesis.

# 2 Related work

## 2.1 Fixed-Length Video Generation Paradigms

Most mainstream video generation models are designed around a predefined temporal support. In diffusion-based video generation, the number of frames is typically determined by the shape of the sampled spatio-temporal noise or latent tensor before denoising begins. Models such as Video Diffusion Models, Imagen Video, Make-A-Video, Video LDM, and recent DiT-based video generators have greatly improved visual fidelity, temporal coherence, and text-conditioned controllability. However, their generation process usually operates over a fixed temporal window. Although the user or sampling procedure may specify different output durations, the model does not make a sequential decision about whether the generated behavior has reached completion. Moreover, because diffusion models operate over continuous pixel or latent representations, they do not naturally provide a categorical endpoint analogous to an end-of-sequence token.

Typical autoregressive and token-based video models are structurally closer to BioVid, but most of them still assume a predefined output format. VideoGPT, HARP, MAGVIT, and related models demonstrate that videos can be represented as latent visual tokens and modeled using Transformer-based architectures for future prediction, reconstruction, or task-conditioned synthesis. Nevertheless, their generation length is generally determined by the training clip length, prediction horizon, mask layout, or task specification. In other words, these models learn how to generate visual content within a given temporal support, rather than how to decide when that support should end. Therefore, while fixed-length diffusion and autoregressive paradigms provide important foundations for video synthesis, they do not directly address adaptive behavioral termination as defined in this work.

## 2.2 Long Video Generation and External Continuation

A second line of related work addresses the limitation of fixed temporal horizons by extending video generation beyond the length used during training. These methods are more closely related to variable-length generation than standard fixed-length models, because they explicitly aim to synthesize longer temporal sequences. However, their central problem is long-horizon continuation rather than semantic termination. In other words, they ask how a model can continue generating coherent video content for more frames, but not how the model should infer that a visually observed behavior has naturally ended.

TATS is a representative autoregressive approach to long video generation. It combines a time-agnostic VQGAN with a time-sensitive Transformer and uses sliding-window generation to synthesize videos substantially longer than the short clips used for training. To reduce quality degradation over long rollouts, TATS introduces a hierarchical design in which sparse latent frames provide global temporal structure and intermediate frames are filled in through interpolation. Similarly, LVDM extends latent diffusion to long video synthesis by using a hierarchical latent video diffusion framework, together with conditional latent perturbation and unconditional guidance to mitigate accumulated degradation during long-range generation [14]. VideoPoet also supports longer video synthesis by autoregressively extending generated content, for example by predicting short video segments conditioned on the previously generated segment. These methods demonstrate that video models can be extended beyond a fixed training horizon while maintaining temporal coherence.

Despite this progress, external continuation is fundamentally different from adaptive behavioral termination. The length of the generated video remains controlled by rollout steps, sliding windows, hierarchical extension stages, generated chunks, or user-specified continuation procedures. The model is asked to keep generating until the external process stops, rather than to decide internally that the action has reached completion. This distinction is especially important for biological behavior generation, where the endpoint of a clip is not merely the end of a sampled horizon but part of the empirical structure of the behavior. BioVid therefore treats duration not as the number of continuation steps to execute, but as a learned endpoint distribution represented by an explicit EOS token.

### 2.3 Phenaki and Prompt-Driven Variable-Length Generation

Phenaki is one of the closest prior works to the notion of variable-length video generation, as it explicitly aims to generate videos of variable duration from open-domain textual descriptions. Its key technical component is C-ViViT, a video tokenizer that compresses videos into discrete tokens using causal attention in time. Unlike tokenizers restricted to fixed-size inputs and outputs, this causal structure allows C-ViViT to encode and decode variable-length videos. This design enables Phenaki to generate videos beyond a single fixed temporal window, making it more relevant to the present work than standard fixed-length video generators.

However, the source of variable length in Phenaki is fundamentally different from adaptive behavioral termination. Phenaki is explicitly designed to generate arbitrarily long videos conditioned on a sequence of prompts, that is, a time-variable text or story that narrates what happens over time. In practice, a user provides a segmented story in which each prompt describes the next stage of the video; the model generates an initial segment, then preserves previously generated tokens and predicts future tokens conditioned on the next prompt. By repeating this process, Phenaki synthesizes a long video whose semantic progression and overall duration follow the provided prompt sequence. Its variable length is therefore a consequence of prompt segmentation and continuation: the video becomes longer because the textual story contains more stages or because the generation process is externally extended.

This mechanism differs from BioVid's definition of adaptive-length behavioral generation. In Phenaki, the temporal extent is governed by the prompt sequence, the story design, and the continuation procedure, rather than by the visual content of the generated behavior. Although Phenaki uses discrete video tokens and temporal causality, its termination is dictated by the exhaustion of the input prompt sequence rather than by any learned signal indicating that a depicted behavior has reached its natural endpoint. Its objective is to generate coherent open-domain videos under time-varying textual control, not to learn the empirical duration distribution of a behavior class from visual data alone. Phenaki thus represents an important form of prompt-driven variable-length video generation, but it does not address the problem of data-driven semantic termination studied in this work.

### 2.4 TV2TV and Interleaved Text-Video Planning

TV2TV represents another recent direction that appears related to adaptive-length generation, because it allows the generation process to dynamically alternate between textual reasoning and video synthesis. Instead of treating text as only an initial conditioning signal, TV2TV decomposes video generation into an interleaved sequence of text segments and video frame chunks. At inference time, the model can first generate a textual plan describing what should happen next, then generate a short chunk of video frames conditioned on that plan, and repeat this process over time. This design allows the model to “think in words” before “acting in pixels,” improving controllability and enabling users to intervene through intermediate textual prompts during generation.

The technical formulation of TV2TV is also different from standard token-only autoregressive video generation. It jointly models discrete text tokens and continuous video latents in an interleaved sequence. The text stream is trained with language-modeling objectives, while the video stream is generated through flow matching over temporally segmented video chunks. TV2TV therefore introduces a dynamic schedule over modalities: the model learns when to produce textual planning content and when to synthesize video content. This makes it more adaptive than fixed-window or externally extended video generation, because the generation trajectory is no longer a simple sequence of homogeneous video frames or tokens.

However, the adaptivity in TV2TV is not the same as adaptive behavioral termination. TV2TV's dynamic decision concerns the alternation between planning and rendering: when should the model generate text, and when should it generate video? This is a modality-level or reasoning-level decision, not a semantic endpoint decision for a visual behavior. The video is still produced in chunks, and the role of intermediate language is to guide what should happen next, improve prompt alignment, or enable user intervention. The model is not trained to infer from visual motion alone that an action has naturally reached completion, nor does it learn the duration distribution of a behavior class as an endpoint distribution.

Therefore, TV2TV is closely related to BioVid in that both challenge the idea of a purely fixed video generation process. However, they address different forms of adaptivity. TV2TV uses interleaved language to plan and control an open-ended video generation trajectory, whereas BioVid uses visual autoregression to decide when a specific behavior should stop. In this sense, TV2TV learns adaptive planning over text-video modalities, while BioVid learns semantic termination over visual behavior tokens.

# 3 Preliminaries

## 3.1 VQ-GAN-based Discrete Visual Tokenization

VQ-GAN provides a common visual tokenization paradigm for token-based image and video generation [15]. Instead of modeling pixels directly, VQ-GAN first compresses an image into a lower-resolution discrete latent representation, and a subsequent generative model can then learn the distribution of these discrete visual tokens. This design separates visual reconstruction from sequence modeling: the tokenizer is responsible for preserving perceptually important visual information, while the Transformer models the composition and temporal evolution of the resulting token sequence.

Given an image $x \in \mathbb{R}^{H \times W \times 3}$, a convolutional encoder $E$ maps it to a latent feature map $z = E(x) \in \mathbb{R}^{h \times w \times d}$.Each spatial latent vector is quantized to an entry in a learned codebook $Z = \{z_k\}_{k=1}^{K}$, and a decoder $G$ reconstructs the image from the quantized representation:

$$z = E(x), \qquad zq = q(z), \qquad \hat{x} = G(zq). \tag{1}$$

The quantization operation maps each latent vector $z_{ij}$ to its nearest codebook entry:

$$q(z_{ij}) = \arg\min_{z_k \in Z} \left\| \boldsymbol{z_{ij}} - \boldsymbol{z_k} \right\|_2 \quad (2)$$

The resulting codebook indices form a two-dimensional discrete token map, which can be flattened into a one-dimensional sequence for Transformer-based modeling. Compared with the original VQ-VAE formulation [16], VQ-GAN improves perceptual reconstruction quality by combining reconstruction, perceptual, and adversarial objectives. This enables stronger compression while preserving visually meaningful local structure, making it suitable as a first-stage tokenizer for autoregressive visual generation.

However, because the codebook in VQ-GAN is learned, training may suffer from codebook under-utilization or inactive entries, reducing the effective vocabulary size. This limitation motivates the quantization redesign introduced in the proposed FSQ-R3GAN tokenizer in Section 4.2.

### 3.2 Autoregressive Sequence Modeling

Autoregressive modeling factorizes the probability of a sequence into a product of conditional next-token distributions. Given a discrete sequence $s = (s_1, s_2, \dots, s_L)$, the joint probability can be written as:

$$p(s) = p(s_1) \cdot p(s_2 \mid s_1) \cdot p(s_3 \mid s_1, s_2) \cdot \dots \cdot p(s_L \mid s_1, s_2, \dots, s_{L-1}) \quad (3)$$

Here, $s_1, s_2, \dots, s_L$ are the tokens in the sequence, and each token is predicted using only the tokens that appear before it. In practice, a causal Transformer is commonly used to parameterize these conditional distributions [17]. The causal attention mask ensures that each position can attend only to previous positions, so the model learns to predict the next token from the history available so far.

This formulation is especially suitable for discrete visual tokens because image or video token maps can be flattened into one-dimensional sequences and modeled in the same way as language tokens. During training, the model receives a prefix of tokens and is optimized to predict the next token at each position. During inference, generation proceeds sequentially: each sampled token is appended to the context and used to condition subsequent predictions.

A key advantage of autoregressive modeling is that special symbols can be incorporated into the same vocabulary as ordinary data tokens. For example, an End-of-Sequence (EOS) token can be appended to the end of a training sequence. The model then learns not only which visual token should come next, but also when the sequence should terminate. At inference time, generation can stop when EOS is emitted, making sequence length a learned outcome of the next-token prediction process rather than a separately prescribed parameter.

# 4 Methodology

## 4.1 Overview

BioVid is a two-stage data-driven framework for adaptive-length biological behavior video generation. The first stage converts each video into a frame-wise sequence of discrete visual tokens using a 2D-encode/3D-decode FSQ-R3GAN tokenizer. The two-dimensional encoder preserves single-frame token representations suitable for autoregressive modeling, while the temporally inflated three-dimensional decoder improves reconstruction consistency across frames. The second stage trains a causal Transformer over this token sequence with an explicit End-of-Sequence (EOS) token appended to each training clip. At inference time, the model is conditioned on the first frame and autoregressively predicts visual tokens until EOS is emitted, as illustrated in Fig. 1. This design keeps the generation process discrete, frame-wise, and self-terminating, allowing video length to be learned from data rather than prescribed as a fixed frame horizon.

## 4.2 2D3D-FSQ-R3GAN tokenizer

We refer to the proposed tokenizer as 2D3D-FSQ-R3GAN, an asymmetric discrete visual tokenizer designed for autoregressive biological behavior generation. It combines finite scalar quantization, an R3GAN-based autoencoding backbone, and a 2D-encode/3D-decode architecture. The goal of this tokenizer is not only to reconstruct frames with high perceptual quality, but also to produce a frame-wise discrete token

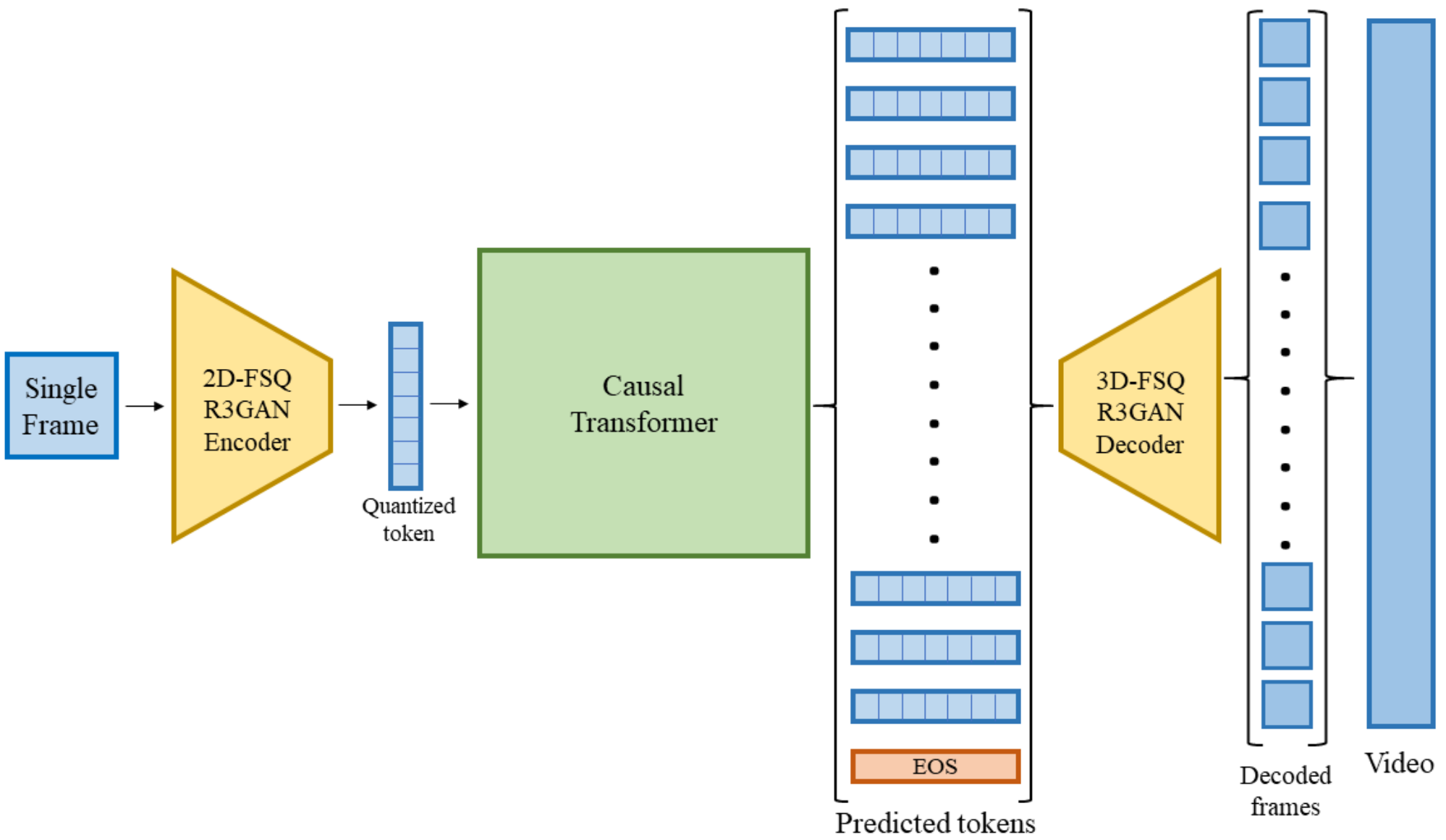

Fig. 1. The inference pipeline of BioVid.

space that remains compatible with causal next-token prediction and EOS-based termination.

**Replacing VQ with FSQ for Stable Discrete Tokenization.** Starting from the VQ-GAN tokenization paradigm, we replace learned vector quantization with Finite Scalar Quantization (FSQ) [18]. In conventional VQ-based tokenizers, each encoder output vector is assigned to the nearest entry in a learned codebook. Although this produces discrete visual tokens, the learned codebook can suffer from under-utilization or inactive entries, especially when the effective data diversity is limited. This instability is undesirable for autoregressive modeling, because the Transformer learns a categorical distribution over the tokenizer vocabulary. In our early experiments, we observed this issue even after reducing the codebook size from the commonly used 1024 entries to 256 entries. Despite the smaller vocabulary, only around 60% of the codebook entries were actively utilized during training, indicating that a substantial portion of the learned codebook remained ineffective.

FSQ avoids this issue by quantizing each latent channel independently into a fixed set of scalar levels. Given a latent vector $z = (z_1, \ldots, z_D)$, each dimension is discretized separately according to

$$q_d(z_d) = arg \min_{l \in \mathcal{L}_d} |z_d - l| \tag{4}$$

where $\mathcal{L}_d$ denotes the predefined set of scalar quantization levels for the $d$-th channel. The quantized latent representation is then obtained by applying scalar quantization independently across all dimensions,

$$\mathrm{q(z)} = (q_1(z_1), q_2(z_2), \ldots, q_D(z_D)). \tag{5}$$

The resulting implicit codebook is formed by the Cartesian product of all scalar levels across dimensions, yielding a fixed vocabulary without requiring a learned embedding table. As a result, the tokenizer obtains a stable finite vocabulary without requiring learned codebook updates, commitment losses, or codebook-reinitialization heuristics. In BioVid, this provides a reliable discrete token space for learning visual token transitions and EOS prediction.

**Adapting R3GAN into a Visual Tokenizer.** We further adapt R3GAN into an autoencoding tokenizer backbone [19]. R3GAN was originally proposed as a modernized GAN framework with a regularized relativistic adversarial objective for stable image generation [20]. Unlike the original formulation, which directly maps random noise to images, we redesign the generator as a decoder and construct a corresponding encoder as its architectural mirror to form an autoencoding architecture.

Specifically, the encoder first compresses each input frame into a low-resolution latent representation, which is then discretized by the FSQ module into visual token indices. The decoder reconstructs the original frame or video from the corresponding quantized embeddings, while the discriminator remains responsible for providing adversarial supervision during training. This modification transforms R3GAN from a pure image generator into a discrete visual tokenizer capable of producing compact token sequences for autoregressive modeling. Following the two-stage tokenization paradigm established by VQ-GAN, our tokenizer separates visual compression from sequence modeling: the encoder-quantizer-decoder architecture learns a discrete visual vocabulary, while the downstream Transformer models the resulting token sequences. However, instead of adopting the original VQ-GAN architecture and training objective, we replace its vector-quantized autoencoder with an FSQ-based R3GAN backbone to obtain more stable quantization and stronger reconstruction quality.

Training follows the core design of R3GAN, which combines a relativistic adversarial objective with R1 and R2 regularization [21]. Let $D(x)$ denote the discriminator output for a real sample $x$, and $D(\hat{x})$ denote the output for a reconstructed sample $\hat{x}$. Rather than evaluating real and reconstructed samples independently, R3GAN compares them through their relative discriminator scores. The discriminator objective additionally incorporates R1 and R2 gradient regularization on real and reconstructed samples, respectively. These regularization terms penalize excessive sensitivity of the discriminator to small input perturbations, encouraging smoother decision boundaries and reducing training instability. As a result, the adversarial optimization process becomes more robust, helping to prevent gradient explosions or oscillatory behavior and leading to more stable convergence during tokenizer training. The generator and discriminator losses are defined as

$$\mathcal{L}_G^{adv} = \mathbb{E}_{x,\hat{x}}[softplus(D(x) - D(\hat{x})], \tag{6}$$

$$\mathcal{L}_D^{adv} = \mathbb{E}_{x,\hat{x}}[softplus(D(\hat{x}) - D(x)] + \frac{\gamma_1}{2} R_1 + \frac{\gamma_2}{2} R_2. \tag{7}$$

By optimizing these relativistic objectives, the discriminator learns to assign higher scores to real samples than reconstructed samples, while the generator is encouraged to produce reconstructions that are indistinguishable from real data in terms of their relative discriminator scores. The R1 and R2 regularization terms further constrain discriminator gradients on real and reconstructed samples, reducing training instability and improving convergence.

Unlike VQ-based tokenizers, FSQ does not require a learned codebook, commitment loss, or codebook update mechanism. Consequently, the tokenizer is optimized using only reconstruction and adversarial objectives. The reconstruction term combines a pixel reconstruction loss and a perceptual LPIPS loss [22],

$$\mathcal{L}_{rec} = \lambda_1 \mathcal{L}_1 + \lambda_{LPIPS} \mathcal{L}_{LPIPS} \tag{8}$$

where $\mathcal{L}_1$ measures pixel-level reconstruction error and $\mathcal{L}_{LPIPS}$ encourages perceptual similarity between reconstructed and ground-truth frames. The overall tokenizer objective is then given by:

$$\mathcal{L}_{tok} = \lambda_{rec} \mathcal{L}_{rec} + s(t) \lambda_{adv} \mathcal{L}_G^{adv} \tag{9}$$

where $s(t)$ is an adversarial gate that delays adversarial training during the early stage of optimization. In our implementation, $s(t) = 0$ for the first 100 kimg and $s(t) = 1$ thereafter. The discriminator is optimized separately using Eq. (7).

**Asymmetric 2D Encoding and 3D Decoding.** A central requirement of BioVid is frame-level termination. Since the goal is to match the natural duration distribution of behavioral events, the generator must be able to stop at any frame boundary rather than only at a fixed temporal interval. Many video tokenizers adopt three-dimensional architectures with temporal downsampling, where several consecutive frames are compressed into a smaller number of latent time steps. This design is effective for reducing sequence length and improving spatio-temporal compression, but it also changes the temporal granularity of the generated representation. If one latent time step corresponds to multiple frames, the autoregressive model effectively generates video in multi-frame units, making it difficult to emit an EOS token at an arbitrary frame boundary. For this reason, our initial design avoided temporal downsampling and used a purely frame-wise two-dimensional tokenizer, so that each group of visual tokens corresponds to exactly one frame.

This frame-wise 2D design provides two important advantages. First, it preserves the single-frame token granularity required for EOS-based adaptive-length generation. The Transformer can generate tokens frame by frame and decide whether to terminate after any completed frame. Second, because each frame is encoded and reconstructed independently, the tokenizer can focus on high-quality spatial reconstruction, which leads to stronger single-frame fidelity. However, this design also exposes a major weakness when the reconstructed frames are viewed as a video. Since each frame is decoded independently using only 2D convolutions, temporal information cannot flow across adjacent frames during reconstruction. As a result, small frame-to-frame reconstruction differences accumulate visually as flickering, reducing temporal consistency even when individual frames are sharp.

To address this limitation, we introduce an asymmetric 2D-encode/3D-decode architecture. The encoder remains two-dimensional and processes each frame independently using 2D convolutions, preserving single-frame visual tokens that are compatible with causal autoregressive modeling. In contrast, the decoder is replaced with a temporally aware three-dimensional decoder composed of 3D convolutions, which reconstructs a video clip from the generated frame-wise token sequence while allowing information to propagate across neighboring frames. This design keeps the token sequence seen by the Transformer unchanged: the model is still trained on frame-wise tokens produced by the 2D encoder, and EOS can still be predicted at any frame boundary. The temporal modeling is introduced only at the reconstruction stage, where it can smooth frame-to-frame variations without altering the autoregressive generation granularity.

The 3D decoder is initialized from the trained 2D decoder through temporal inflation, following the 2D-to-3D inflation strategy used in MAGVIT. In the original MAGVIT design, temporal inflation is combined with temporal downsampling, where multiple consecutive frames are compressed into a reduced number of latent time steps. While this design improves compression efficiency and enables joint spatio-temporal modeling, it also changes the temporal granularity of the representation, making each latent token correspond to multiple frames. In contrast, to preserve frame-level token granularity for autoregressive generation, we remove the temporal downsampling mechanism and retain a one-to-one correspondence between latent tokens and input frames. Specifically, each two-dimensional convolution kernel is expanded into a three-dimensional convolution kernel by adding a temporal dimension. The pretrained 2D weights are placed in the center temporal slice of the 3D kernel, while the remaining temporal slices are initialized to zero. Parameters that do not depend on the convolutional temporal dimension are copied directly from the 2D decoder when their shapes match. After inflation, the decoder is fine-tuned on video clips so that the newly introduced temporal slices learn to integrate information from neighboring frames. This initialization preserves the spatial reconstruction ability learned by the 2D decoder while enabling the decoder to acquire temporal smoothing capability from video data, without introducing temporal downsampling that would compromise frame-level generation control.

Unlike fully three-dimensional tokenizers, we do not inflate the entire encoder–decoder architecture. A 3D encoder would use 3D convolutions to aggregate information from both past and future frames, producing tokens that contain bidirectional temporal context. While this is beneficial for reconstruction, it is less compatible with causal autoregressive generation where future frames are unavailable at inference time. Instead, BioVid uses the 2D encoder to define the autoregressive token space and the inflated 3D decoder only to improve reconstruction. This asymmetric design retains frame-level token granularity for adaptive EOS prediction while leveraging temporal context during decoding to reduce flickering and improve video consistency. As shown in the ablation study in Section 5.1, this 2D3D design provides the best trade-off among length distribution matching, temporal coherence, and action recognition quality.

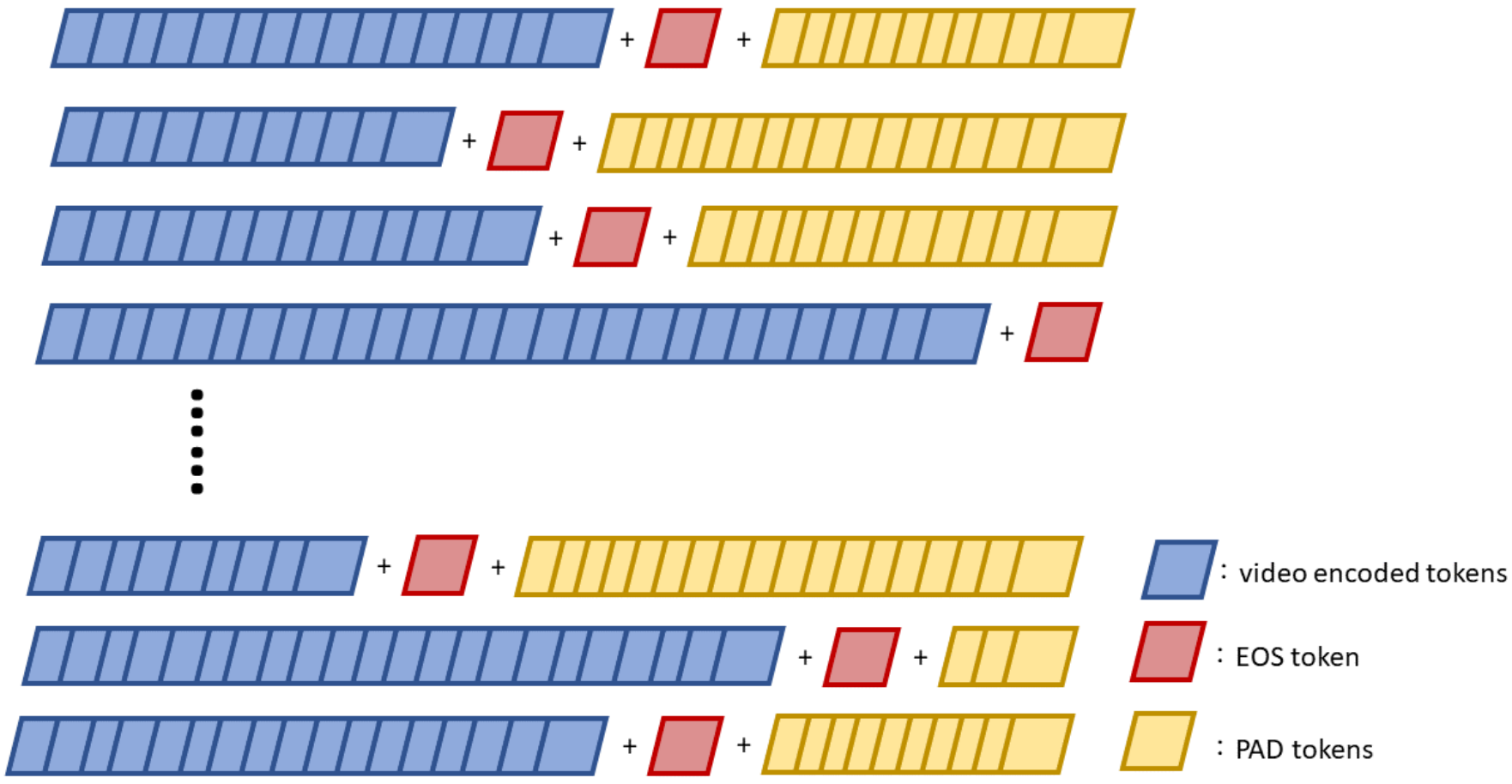

Fig. 2. **EOS/PAD Scheme for Variable-Length Video Tokens.** Each video is encoded into a variable-length sequence of visual tokens. An EOS token is appended after the last visual token to indicate the stopping point, and PAD tokens are added to align all sequences to the maximum length.

### 4.3 Variable-Length Autoregressive Transformer

The decoded token grids are rasterized into a one-dimensional sequence and modeled by a causal decoder-only Transformer with approximately 66M trainable parameters. The central contribution of this stage is a mechanism that allows the model to determine video length from data rather than from a fixed architectural horizon.

**EOS mechanism for variable-length generation.** We augment the FSQ vocabulary with two special symbols — EOS and PAD. Each preprocessed frames is encoded independently by the trained 2D3D-FSQ-R3GAN tokenizer into a 16×16 token grid, or 256 visual tokens per frame. For a clip with T frames, the frame-level grids are flattened in raster order and concatenated along time to form a visual-token sequence of length 256T. We append one EOS token immediately after the final visual token of the clip. Thus, the endpoint supervision is derived directly from the empirical end of each preprocessed clip, rather than requiring additional manual stop annotations. The EOS token is heavily up-weighted in the loss (eos_loss_weight = 10.0). Because EOS appears only once per sequence yet governs the entire length prediction, an unweighted objective would underemphasize it and the model would fail to learn when to stop; the 10× weighting makes EOS prediction dependable. In contrast, PAD tokens serve a purely structural role: during batching, variable-length sequences are padded with PAD tokens to the length of the longest sequence in the dataset, as shown in Fig. 2; PAD is used only for tensor alignment and is masked out of the loss entirely.

The Transformer is trained with shifted next-token prediction: the input sequence contains all tokens except the last one, and the target sequence is shifted by one position. As a result, the model learns both visual-token continuation and when to emit EOS. At inference time, generation starts from the first-frame prompt and terminates naturally when EOS is generated, instead of relying on a fixed frame horizon.

**3D rotary positional encoding.** Rather than a conventional 1D positional embedding, we apply 3D RoPE to the rasterized token sequence [23]. The temporal, height, and width axes each receive an independent rotary embedding, which are then concatenated. This makes positional encoding aware of the underlying 3D structure of the video rather than treating the tokens as a spatially meaningless 1D stream, allowing the model to implicitly infer which frame, row, and column a token belongs to — beneficial for both spatial coherence and temporal modeling.

### 4.4 Training and Inference Strategy

**Human-centered preprocessing.** Our initial experiments showed that directly using the standard NTU RGB+D center crop was insufficient for generation. The actor's location and apparent scale still varied across clips, causing the same body part to appear in different spatial regions and forcing the model to spend capacity on crop-induced misalignment. We therefore use YOLO to detect the human subject in each clip, smooth the detected bounding boxes over time, crop around the actor, and resize the result to 256×256 [24]. This produces a temporally ordered frame directory for each clip and canonicalizes the actor's position and scale, which substantially improves the consistency of both tokenizer reconstruction and autoregressive generation.

**ROI-aware loss weighting.** Although the human-centered crop places the actor consistently in the frame, it also introduces a distributional imbalance: the background regions remain nearly static across frames while the human body undergoes rapid and complex motion. Left unconstrained, the model tends to minimize loss by over-fitting the easy-to-learn static background, leaving the semantically critical human body region insufficiently learned. To counteract this, we apply a spatially weighted loss that forces the model to prioritize the human body region. Since the actor is consistently near the center after preprocessing, we up-weight the central human-body region and down-weight peripheral background regions. In FSQ-R3GAN, this spatial mask is applied to the reconstruction objective so that errors on the actor contribute more strongly than errors on mostly static background. In the Transformer, the same idea is applied to the 16×16 token grid, assigning higher weights to central visual tokens and lower weights to edge tokens. The mask is normalized to have unit mean, keeping the average loss scale comparable to the unweighted objective while reallocating model capacity toward the action-relevant region.

**Two-stage pretrain–finetune.** In early experiments, we trained the Transformer directly on a single target behavior class. The model rapidly memorized the training set and exhibited a critical failure mode at inference: when conditioned on a first frame outside the training distribution, the model would immediately emit EOS, collapsing the generated sequence to a single token. This behavior indicated that the model had learned to associate unseen first frames with termination rather than generation, a consequence of insufficient generalization from the small single-class dataset. To resolve this, we adopted a two-stage training strategy. We first pretrain the Transformer on the full NTU RGB+D dataset spanning actions A001–A049 (single-person action classes), with an action embedding supplied as an additional conditioning signal. This stage teaches the model a broad token-level language of human motion, enabling it to generate coherent sequences conditioned on arbitrary first frames. We then fine-tune on the target class A001, specializing the model to the target behavior's length distribution and semantic structure. After fine-tuning, the premature EOS collapse was eliminated: the model consistently produces complete behavioral sequences even when conditioned on held-out first frames.

**Generation diversity.** Having established training stability through the two-stage strategy, we further explored techniques to improve generation diversity and robustness. We apply scheduled sampling during fine-tuning, stochastically substituting the model's own predictions for ground-truth tokens with probability 0.2 [25]. Naive teacher forcing trains the model exclusively on ground-truth context, whereas at inference the model must condition on its own outputs — a distributional mismatch known as exposure bias that causes errors to compound over long sequences. Scheduled sampling bridges this gap by exposing the model to its own predictions during training. At inference, we apply classifier-free guidance (CFG): a MASK token is introduced and used to randomly replace the first-frame tokens with probability 0.1 during training, producing unconditional examples under the same model [26]. At inference, conditional and unconditional logits are interpolated as logits_guided = logits_cond + cfg_scale × (logits_cond − logits_uncond). Tokens are then sampled using temperature scaling and top-k filtering, where temperature controls the sharpness of the predicted distribution and top-k restricts sampling to the k most probable tokens [27]. The configuration cfg_scale = 1.5, temperature = 1.0, and top-k = 20 represents the optimal setting identified through systematic parameter search.

# 5 Experimental Results

## 5.1 Experimental Setup

**Datasets.** All experiments are conducted on the NTU RGB+D dataset (Shahroudy et al., 2016), focusing on the drinking behavior class (A001). Prior to training, all clips undergo the human-centered preprocessing described in Section 4.4: YOLO-based person detection, temporally smoothed bounding box cropping, and resizing to 256×256. For BioVid, we apply a stride-3 frame subsampling strategy to ensure sufficient inter-frame variation, as adjacent frames in the original clips are often nearly identical. For fixed-length baseline models, we apply uniform frame subsampling to match the target sequence length. The dataset is split as follows: pretraining uses 41,846 clips spanning all single-person action classes (A001–A049), fine-tuning uses 854 clips from A001, and evaluation uses 94 held-out test clips from A001.

**Baselines and training procedure.** Due to the domain-specific nature of the NTU RGB+D A001 setting, no publicly available model checkpoint is directly applicable to our evaluation protocol. We therefore compare BioVid with representative baselines from several major video generation paradigms: VideoGPT and TATS for discrete autoregressive video generation, MaskGIT-FSQ for bidirectional masked-token generation, Wan2.1 for latent diffusion-based image-to-video generation, and NOVA for non-quantized autoregressive diffusion. Since MAGVIT does not provide an official PyTorch implementation directly compatible with our experimental pipeline, we additionally implement a simplified PyTorch masked Transformer baseline over FSQ-R3GAN tokens [28]. This baseline, denoted as MaskGIT-FSQ, is used to evaluate the effect of masked prediction and iterative unmasking, rather than to reproduce the full MAGVIT system. For token-based models trained from scratch, we match the Transformer capacity, input resolution, batch size, and optimization settings to BioVid whenever permitted by the corresponding architecture. Training follows the two-stage procedure described in Section 4.4: pretraining on actions A001–A049, followed by fine-tuning on A001. MaskGIT-FSQ shares BioVid's FSQ-R3GAN tokenizer, whereas VideoGPT and TATS retain their native 3D tokenizers. Wan2.1 and NOVA use their released general-domain checkpoints and are adapted to the A001 setting through image-to-video fine-tuning. Specifically, Wan2.1 is fine-tuned with rank-32 LoRA [29], while NOVA undergoes full image-to-video fine-tuning. Their original zero-shot checkpoints are also evaluated to assess the effect of domain adaptation. Conditioning is implemented according to each model's native representation. BioVid and MaskGIT-FSQ use the 2D tokens of the first RGB frame as the generation condition, while Wan2.1 and NOVA directly receive the first RGB frame. TATS and VideoGPT use the first temporal latent slice produced by their 3D tokenizers, which summarizes approximately the first four sampled frames and may therefore contain more temporal information than BioVid's single-frame condition.

**Metrics.** Our evaluation is designed around two primary aspects: the duration distribution and the quality of the generated videos. For duration, we report the mean video length to measure average action duration, the coefficient of variation (CV) to quantify relative duration variability, skewness to characterize distributional asymmetry and long or short tails, and the Wasserstein-1 (W-1) distance to measure the overall discrepancy between generated and ground-truth duration distributions. For video quality, we use FID to assess frame-level visual fidelity and FVD to evaluate spatiotemporal realism and temporal consistency [30], [31]. We additionally introduce an Action Score to measure how clearly and consistently the generated motion is recognized as the target behavior. Action Score employs a VideoMAE video classifier fine-tuned on NTU RGB+D [32]. For each generated video, we determine the rank $r_i$ of the target *drink water* class among the classifier outputs. Videos ranked outside the top five receive zero points, while ranks 1-5 receive 16, 8, 4, 2, and 1 points, respectively. The overall score is

$$S_{action} = \sum_{i=1}^{94} \begin{cases} 2^{5-r_i}, & 1 \le r_i \le 5, \\ 0, & r_i > 5. \end{cases}$$

After comparing several weighting schemes, we adopted exponential weighting to emphasize highly ranked predictions while retaining partial credit for Top-5 recognition. We also report the Top-1 and Top-5 target-recognition rates, defined as the percentages of generated videos for which the target action is ranked first or among the five highest-ranked predictions, respectively.

### 5.2 Architecture Validation and Ablation Study

Before comparing BioVid with external baselines, we first validate the key design choices introduced in Section 4. This section evaluates the effects of FSQ tokenization, EOS loss weighting, and the 2D-encode/3D-decode tokenizer design.

We first examine the quantization choice in the proposed tokenizer. As discussed in Section 4.2, replacing the learned VQ codebook with FSQ is expected to improve code utilization and stabilize discrete representation learning. The comparison in Table 1 and Fig. 3 confirms this design choice. Quantitatively, FSQ-R3GAN achieves a lower rFID than VQ-R3GAN, reducing the score from 25.82 to 22.28. Qualitatively, FSQ also produces cleaner reconstructions with fewer local artifacts and better preservation of subject appearance. These results indicate that FSQ provides a more reliable visual token space for the subsequent autoregressive Transformer.

Having established FSQ as the quantization mechanism, we next investigate how temporal information should be incorporated into the tokenizer architecture. As shown in Table 2, our initial design adopted a fully 2D tokenizer to preserve frame-level token granularity and enable EOS-based termination at any frame. Although this design achieved the best frame-level fidelity (FID 39.31) and closely matched the ground-truth duration distribution (W-1 1.24), independent frame decoding produced noticeable temporal flickering, as reflected by an FVD of 42.60 and an Action Score of 736. We therefore investigated a conventional 3D tokenizer with temporal downsampling. However, this design substantially degraded frame-level fidelity (FID 48.81) and caused the generated duration distribution to deviate markedly from the ground truth (W-1 6.30), despite improving FVD to 37.77. Removing temporal downsampling improved the quality metrics, reducing FID and FVD to 47.31 and 36.47 while increasing the Action Score to 825. Nevertheless, despite restoring a one-to-one

Table 1. rFID comparison of VQ and FSQ.

| Quantization | rFID↓ |
|---|---|
| VQ | 25.82 |
| FSQ | **22.28** |

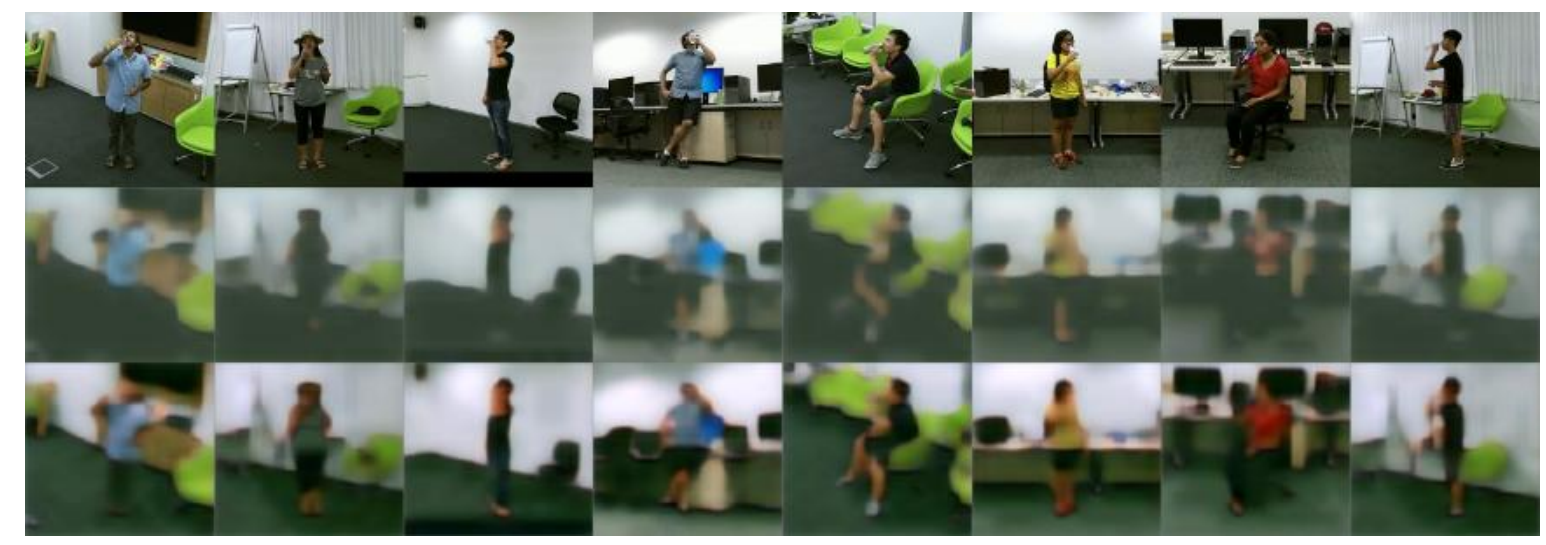

Fig. 3. Qualitative comparison of VQ and FSQ reconstructions. Evaluated at the identical 100k-image training progress. The top, mid, bottom rows are ground truth, VQ baseline, and FSQ, respectively.

Table 2.Ablation study of 2D and 3D tokenizer architectures. TDS denotes temporal downsampling. For Mean, CV, and Skewness, bold values indicate the closest match to the ground-truth distribution; for W-1, bold indicates the lowest distance among the generated models. Action Score is accumulated over 94 test videos.

(a) Length-distribution metrics.

| Tokenizer architecture | Mean | CV (%) | Skewness | W-1↓ |
|---|---|---|---|---|
| Ground truth | 31.48 | 23.86 | 0.70 | 0.00 |
| 2D–2D | **30.68** | **21.56** | **0.31** | **1.24** |
| 2D–3D | **30.68** | **21.56** | **0.31** | **1.24** |
| 3D–3D without TDS | 28.17 | 35.41 | -0.93 | 3.63 |
| 3D–3D with TDS | 25.18 | 48.75 | -0.86 | 6.30 |

(b) quality metrics.

| Tokenizer architecture | FID↓ | FVD↓ | Top-1(%)↑ | Top-5(%)↑ | Action score↑ |
|---|---|---|---|---|---|
| 2D–2D | **39.31** | 42.60 | 39.4 | 66.0 | 736 |
| 2D–3D | 44.89 | **34.01** | 48.9 | **78.7** | **912** |
| 3D–3D without TDS | 47.31 | 36.47 | **50.0** | 69.1 | 825 |
| 3D–3D with TDS | 48.81 | 37.77 | 36.2 | 77.7 | 727 |

correspondence between latent temporal steps and output frames, its W-1 distance remained 3.63, substantially higher than the 1.24 achieved by the fully 2D tokenizer under the same frame-level generation granularity. This observation motivated our asymmetric 2D–3D design, which retains 2D frame-wise tokens for autoregressive modeling and termination while employing a 3D decoder for temporally contextualized reconstruction. The resulting model preserved the W-1 distance of 1.24 while achieving the best FVD of 34.01 and Action Score of 912, providing the most effective balance between adaptive-length modeling and temporal consistency.

We next evaluate the effect of EOS loss weighting, introduced in Section 4.3 to address the sparsity of the EOS token. Since each sequence contains many visual tokens but only a single EOS token, the termination signal can be easily dominated by the visual prediction objective. As shown in Fig. 4, when the EOS weight is set to 1, the model tends to generate overly long sequences, with a mean length of 35.78 frames and a pronounced long-tail distribution extending beyond the ground-truth range, resulting in a Wasserstein-1 distance of 4.32 frames. Increasing the EOS weight significantly improves termination behavior: with a weight of 5, the mean length becomes 30.78 frames, close to the ground-truth mean of 31.48 frames, and the Wasserstein-1 distance decreases to 1.32 frames. Further increasing the weight to 10 yields the best match, with a mean length of 30.68 frames and a Wasserstein-1 distance of 1.24 frames, producing a distribution that closely aligns with the ground-truth without excessive long-tail behavior. These results demonstrate that explicit emphasis on EOS prediction is essential for adaptive-length generation, as insufficient weighting leads to unreliable termination, while a 10× EOS loss weight provides the most stable and accurate stopping behavior in our final BioVid configuration.

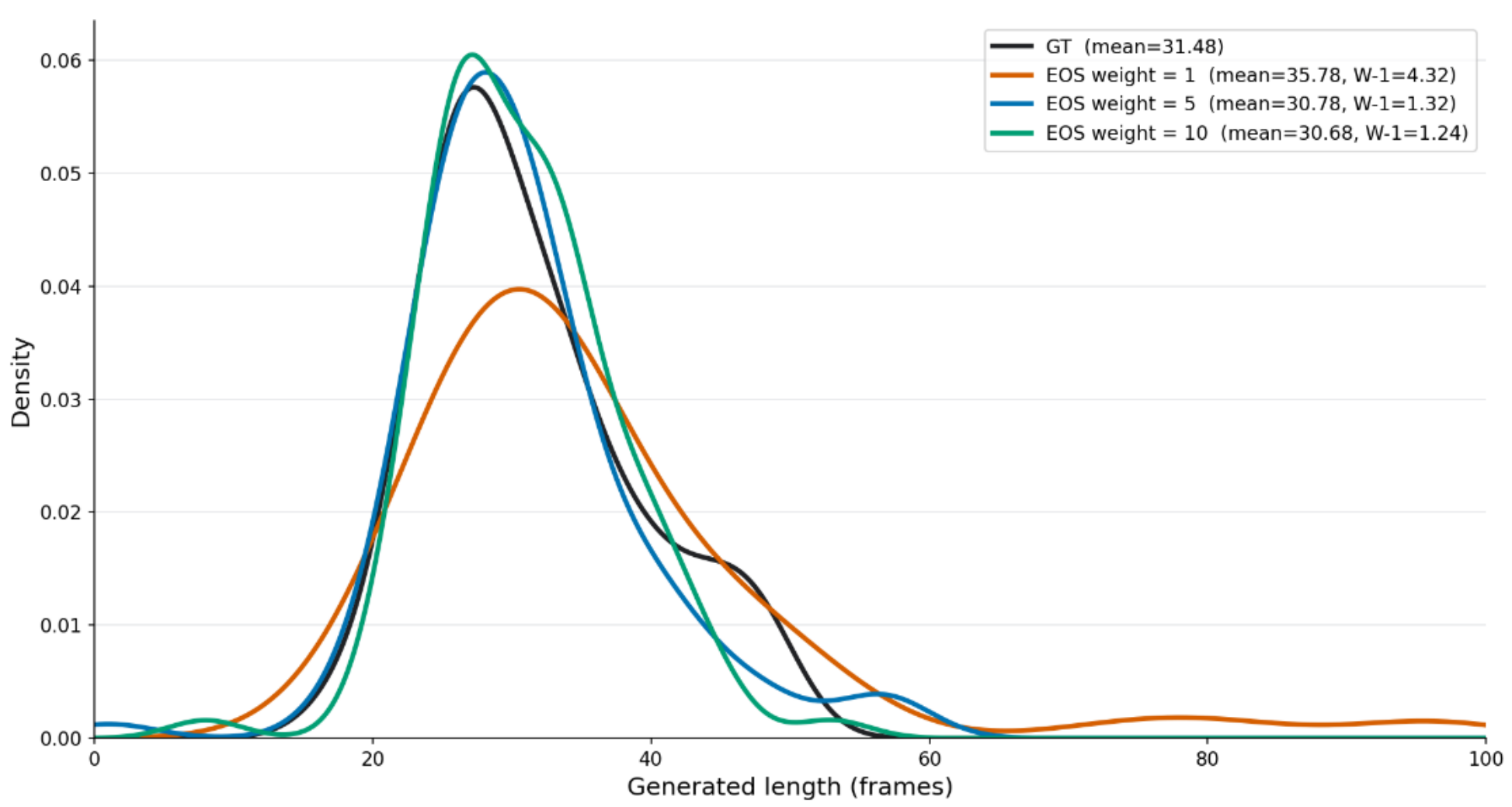

Fig. 4. Length distribution under different EOS loss weights. Each curve represents the kernel density estimate of video lengths for the ground truth or a specific EOS-loss weight.

### 5.3 Adaptive-Length Generation Comparison

We next compare the generated duration distributions between BioVid and the baseline models. Most existing video generation models adopt temporal downsampling or chunk-based generation mechanisms, so their smallest controllable output unit corresponds to multiple frames rather than a single frame. As a result, these models cannot freely terminate at arbitrary frame boundaries. To provide a fair comparison, we configure each baseline to generate the available output length closest to the mean duration of the A001 test distribution. In addition, we include a variant of BioVid without the EOS termination mechanism, where the output length is fixed to 31 frames, to isolate the effect of adaptive termination.

The results are reported in Table 2 and Fig. 5. BioVid achieves the closest match to the ground-truth duration distribution, showing both a similar mean length and a similar distributional shape. In contrast, the baseline models generate videos with predetermined lengths. Their generated duration distributions therefore collapse into delta-like distributions centered at their selected output lengths. Although these lengths are chosen to be close to the dataset mean, they cannot reproduce the natural variability of real behavior durations. This limitation is reflected in their substantially larger Wasserstein-1 distances compared with BioVid. The BioVid variant without EOS similarly produces a fixed-length distribution centered at 31 frames, further confirming that the ability to model duration variability arises from the EOS-based termination mechanism.

We also report the conventional 16-frame setting commonly used by many video generation models. Since the 16-frame configuration yields identical duration statistics across fixed-length models, we use VideoGPT-16 as a representative case. This setting deviates strongly from the real A001 duration distribution, demonstrating that the conventional fixed-window evaluation protocol is unsuitable for measuring adaptive-length biological behavior generation. Overall, these results show that matching the average duration is not sufficient: a model must reproduce the full duration distribution, which BioVid achieves through EOS-based semantic termination.

### 5.4 Video Generation Quality Comparison

We finally evaluate the quality of the generated videos. The purpose of this comparison is to verify whether BioVid can achieve adaptive-length generation while maintaining competitive visual fidelity and behavioral consistency. As shown in Table 4, BioVid achieves the best FID and FVD among models with comparable scale and training conditions, indicating stronger frame-level visual quality and temporal coherence. For the remaining metrics, BioVid maintains competitive performance across all baselines. Notably, VideoGPT achieves a particularly high action score; however, its FID is significantly worse, and the generated videos appear visually blurry. This suggests that the model may be producing outputs that align with the classifier's

Table 3.Generated duration distribution comparison between BioVid and baseline models. For Mean, CV, and Skewness, bold values indicate the closest match to the ground-truth distribution; for W-1, bold indicates the lowest distance among the generated models.

| Model | Mean | CV(%) | Skewness | W-1↓ |
|---|---|---|---|---|
| Ground Truth | 31.48 | 23.86 | 0.70 | 0.00 |
| VideoGPT-16 | 16.00 | 0.00 | N/A | 15.48 |
| VideoGPT | 32.00 | 0.00 | N/A | 6.18 |
| TATS | 32.00 | 0.00 | N/A | 6.18 |
| MaskGIT-FSQ | 32.00 | 0.00 | N/A | 6.18 |
| Wan2.1 | 33.00 | 0.00 | N/A | 6.44 |
| NOVA | 33.00 | 0.00 | N/A | 6.44 |
| BioVid w/o EOS | **31.00** | 0.00 | N/A | 6.05 |
| BioVid (ours) | 30.68 | **21.56** | **0.31** | **1.24** |

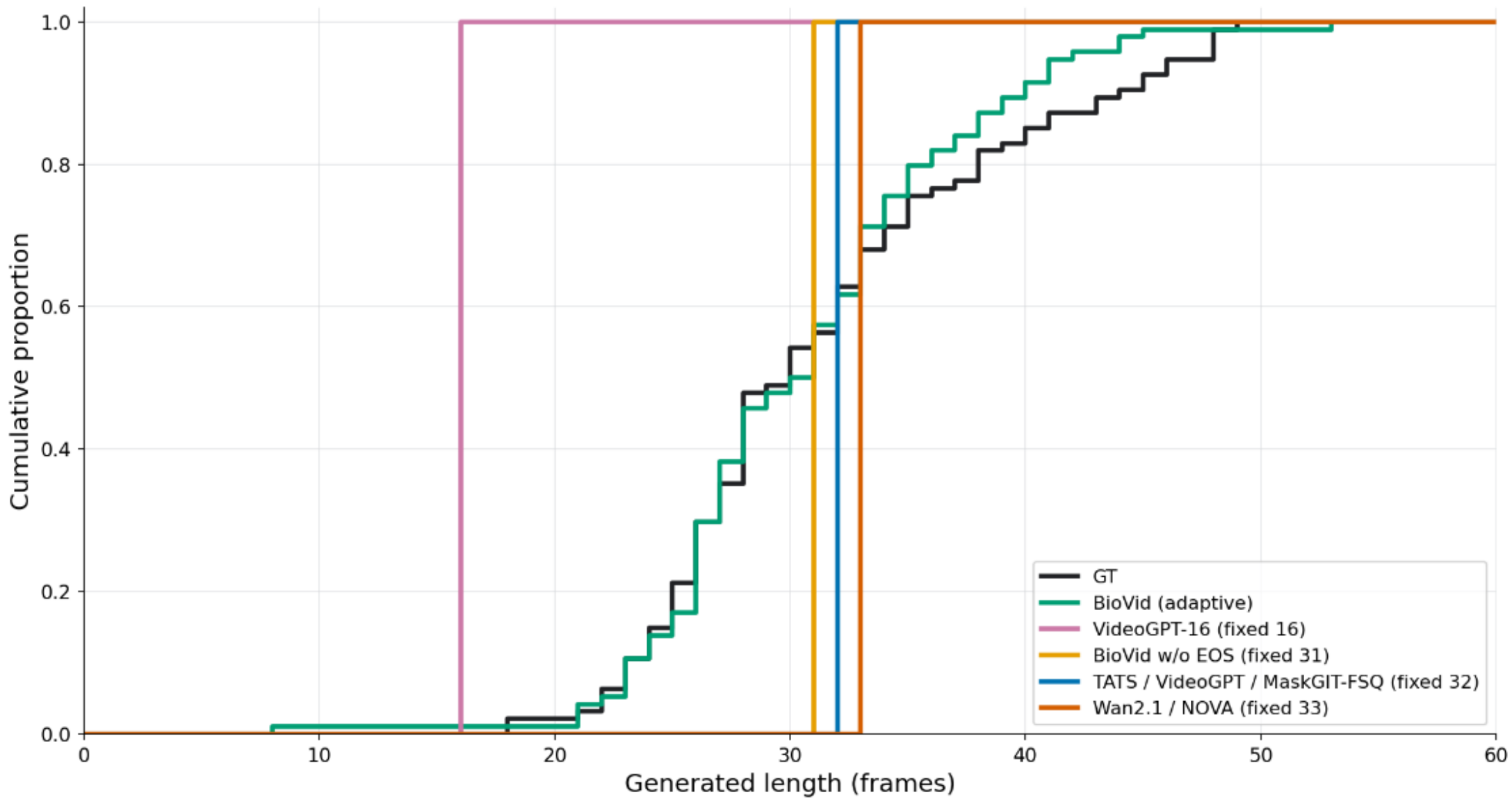

Fig. 5. Empirical cumulative distribution of baseline video lengths. The ECDF compares cumulative video-length distributions. Fixed-length baselines appear as jumps at 16, 31, 32, or 33 frames, with models sharing the same output length represented by a single line.

decision patterns rather than faithfully preserving realistic visual details.

We also include Wan2.1 and NOVA as large-scale pretrained baselines for reference. While these models achieve state-of-the-art performance, their superiority is largely built upon significantly larger parameter scales. Moreover, despite their high visual quality, they still lack the element of speed variability. However, these models are not directly comparable to BioVid, because they rely on substantially different generation architectures, broad general-domain pretraining, and parameter counts that are tens to hundreds of times larger than BioVid. Therefore, their results should be interpreted as reference points for large-scale pretrained video generation rather than as direct competitors under the same experimental setting.

In addition to fine-tuned results, we report the performance of the original pretrained checkpoints without A001-specific adaptation. For these large-scale models, this setting simulates practical scenarios where full fine-tuning may be computationally prohibitive, or where only a closed-source or fixed pretrained model is available. The comparison shows that domain adaptation is important for biological behavior generation: without fine-tuning, large pretrained models exhibit a substantial drop in performance, failing to preserve the target action distribution and behavioral semantics, and performing significantly worse than models adapted to the A001 setting.

## 6 Limitations and Future Work

BioVid is currently designed for generating short, well-defined biological actions. This setting is suitable for our target task, where the generated video only needs to capture a complete and recognizable behavior. However, the model has not yet been extended to long-duration video generation. In future work, BioVid could be combined with long-video generation techniques inspired by TATS, such as hierarchical temporal modeling or segment-wise generation, to support longer and more complex biological

Table 4. Generated video quality comparison between BioVid and baseline models. Baselines are grouped by model scale.

| Model | FID↓ | FVD↓ | Top-1 (%)↑ | Top-5 (%)↑ | Action↑ |
|---|---|---|---|---|---|
| VideoGPT | 96.39 | 45.61 | 76.6 | 97.9 | 1243 |
| TATS | 75.16 | 43.75 | 55.3 | 87.2 | 1001 |
| MaskGIT-FSQ | 45.39 | 39.97 | 0.0 | 14.9 | 60 |
| Wan2.1-zero shot | 97.79 | 88.35 | 4.3 | 14.9 | 111 |
| Wan2.1 | **36.87** | **21.29** | **89.4** | **100.0** | **1412** |
| NOVA-zero shot | 75.35 | 52.78 | 1.1 | 94.7 | 208 |
| NOVA | 50.56 | 28.52 | 75.5 | 94.7 | 1241 |
| BioVid (ours) | 44.89 | 34.01 | 48.9 | 78.7 | 912 |

behavior sequences.

## 7 Conclusion

In this work, we proposed BioVid, a data-driven autoregressive framework for adaptive-length biological behavior video generation. Unlike conventional video generation models that rely on fixed temporal windows, external continuation procedures, or prompt-specified durations, BioVid treats action duration as an intrinsic semantic property of the behavior. By representing videos as frame-wise discrete visual tokens and appending an explicit EOS token to each training sequence, BioVid learns not only how visual motion evolves, but also when the behavior should terminate.

To support this objective, BioVid introduces a 2D-encode/3D-decode FSQ-R3GAN tokenizer. The 2D encoder preserves frame-level token granularity required for causal autoregressive modeling and EOS-based termination, while the temporally inflated 3D decoder improves video consistency by incorporating temporal context during reconstruction. Ablation studies show that FSQ improves tokenizer reconstruction quality over VQ, EOS loss weighting is essential for reliable termination, and the proposed 2D3D tokenizer provides the best trade-off between duration modeling, temporal coherence, and behavioral recognizability.

Experiments on the NTU RGB+D A001 drinking action demonstrate that BioVid closely reproduces the natural duration distribution of real behavioral clips. On 94 held-out test videos, BioVid achieves a Wasserstein-1 distance of 1.24 frames from the ground-truth length distribution, whereas fixed-length baselines remain substantially worse even when configured near the dataset mean. This confirms that matching only the average duration is insufficient; biologically meaningful generation requires modeling the full distribution of action endpoints. In addition, BioVid maintains competitive video quality and achieves the best FID and FVD among models with comparable scale and training conditions.

Overall, BioVid establishes adaptive-length generation as a distinct requirement for biological behavior synthesis. The results show that semantic termination can be learned directly from visual data through discrete autoregressive modeling, enabling generated videos to end according to the learned behavioral distribution rather than a manually prescribed frame count. Future work may extend this framework to longer and more complex behavioral sequences, multi-action settings, and broader biological domains where temporal structure is itself part of the scientific signal.